# A Study on the Lombard Effect in Telepresence Robotics

Ambre Davat[1,2], Gang Feng[1], Véronique Aubergé[2]

[1] GIPSA-lab, Univ. Grenoble Alpes, CNRS, Grenoble INP*, Grenoble, France
[2] LIG, Univ. Grenoble Alpes, CNRS, Grenoble INP*, Grenoble, France
* Institute of Engineering Univ. Grenoble Alpes

Corresponding Author:
Ambre Davat[1,2]
11 rue des Mathématiques, Saint Martin d'Héres, 38400, France
Email address: ambre.davat@gipsa-lab.fr

## Abstract
In this study, we present a new experiment in order to study the Lombard effect in telepresence robotics. In this experiment, one person talks with a robot controled remotely by someone in a different room. The remote pilot (R) is immersed in both environments, while the local interlocutor (L) interacts directly with the robot. In this context, the position of the noise source, in the remote or in the local room, may modify the subjects' voice adaptations. In order to study in details this phenomenon, we propose four particular conditions: no added noise, noise in room R heard only by R, virtual noise in room L heard only by R, and noise in room L heard by both R and L. We measured the variations of maximum intensity in order to quantify the Lombard effect. Our results show that there is indeed a modification of voice intensity in all noisy conditions. However, the amplitude of this modification varies depending on the condition.

## Introduction

When social entities are interacting, they automatically adapt their behavior to adverse conditions which could jeopardize their communication. In particular, vocal modifications due to noise are known as the Lombard effect. This phenomenon was first documented in humans (Lombard, 1911), then in other animal species, especially birds and mammals (Zollinger & Brumm, 2011). Moreover, the Lombard effect has implications for the design of virtual agents and social robots. Indeed, it can impair speech recognition systems, which are generally based on speech corpus recorded in quiet conditions (Hanson & Applebaum, 1990; Junqua, 1993). Implementing the Lombard effect in dialog systems would also increase their adaptation to noisy environment in a biomimetic way.

The Lombard effect consists mainly in an increase of speech intensity. It is an automatic phenomenon, which can only partly be controlled by the speaker (Pick et al., 1989). It varies with the type of noise and differs greatly from one individual to another. It also depends on the speaker's involvement. Indeed, there is some evidence showing that the Lombard effect is stronger during more interactive tasks: It increases during story-telling vs. labelling (Amazi Deborah K. & Garber Sharon R., 1982), communication with another vs. reading (Junqua, Fincke & Field, 1999), or vs. self-talk (Garnier, Henrich & Dubois, 2010). The sound immersion techniques used in such experiments are also important, as speech modifications may be stronger when noise is played through headphones than over loudspeakers (Garnier, Henrich & Dubois, 2010). Furthermore, the intensity is well known as one of the prosodic parameters much implied for producing different socio-affective attitudes (authority, surprise etc.) in many languages (Aubergé, 2015). In face to face interaction, the intensity variations due to the Lombard effect are almost never confused with those having a socio-affective meaning. However, in the case of remote interaction, an ill-formed Lombard effect could be perceivably confused with social affect cues (Aubergé, 2017).

The paper will focus on a study concerning the Lombard effect in telepresence robotics. The specificity of this context is that all interlocutors are not present in the same room: one of them interacts remotely through a robot, which embodies its pilot in the "local" space. It is therefore an asymmetric system because the pilot needs to be acoustically immersed in both remote and local environment, while the interlocutors talk directly with the robot in the local environment. Because the Lombard effect happens automatically, it should affect the pilot's voice regardless of where the noise comes from. On the opposite, if the noise occurs in the remote space, the interlocutors cannot hear it as loud as the pilot, potentially not at all. However, they can hear the pilot's vocal adaptations, which could modify their behavior. In this paper, we propose an experiment in order to study these assumptions and present some results.

## Materials & Methods

**Methodology**

In order to realize this study, we designed an experiment involving two subjects separated in two different rooms as shown on Figure 1. One of them played the role of the remote pilot (R), and the other one was the local interlocutor (L), interacting with the telepresence robot. Several configurations were tested in order to isolate the features of the Lombard effect in remote communication.

**Test conditions**
A: quiet in both rooms
B: noise in the remote space
C: "virtual noise" in the local space
D: noise in the local space

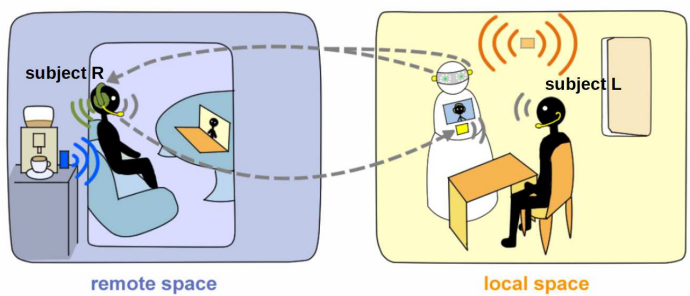

**Figure 1.** *Experimental set-up.*

The **condition A** was used in order to have a reference with no added noise. In all other conditions, R heard a noise, but this noise was not always audible by L. In **condition B**, a noise was played in R's room. As R voice was recorded with a headworn microphone, this noise was largely attenuated for the other subject, thus there was no need for R to speak louder to improve speech intelligibility. On the contrary, even if L did not hear the noise in this condition, L could need to speak louder in order to be understood by R. In **condition D**, a noise was played in the local space, and both subjects could hear it. This case is close to the standard communication with noise, where interlocutors are sharing the same environment. Finally, in **condition C**, a pre-recorded noise was injected in the headphones of R, in order to make R believe that this sound was existing in the local space and was heard by L.

**Procedure**
The experiment was introduced to the subjects as a test of the visual interface of a telepresence robot. The subject L was sitting in front of the robot and asked a list of simple questions to the subject R, who was the pilot of the robot. These questions were designed in order to be very simple, and trigger a finite set of answers. Examples of these questions (translated in English) are: "Which is the color of the grass?", "How many legs does a horse have?", "How much are 2 x 2?". Every 10 questions, there was a special instruction, requesting R to handle the robot via the interface. We pretended that we were measuring the answering time, in order to quantify the accessibility of the pilot interface. The subjects were informed that the test has to be done without a break. Indeed, during our first test, the subject L stopped speaking when the noise D was playing. The three different noises were regularly triggered by an experimenter.

To prevent bias in the experimental results, we wanted to nudge the subjects into thinking that noise occurrences had no link with the experiment. The noise sources were therefore hidden and diffused occasionally. We used pre-recorded realistic noises of a coffee machine and a drill. A ladder and a tool box were placed in the corridor at the entrance of the platform to suggest that there were building works in progress. Before the debriefing, most of the subjects believed that the noises were incidental, and none of them guessed that the aim of the experiment was to measure their vocal adaptation to noise. This confirms that our scenario was credible. Fulfilling the list of 140 questions took approximately 15 minutes long.

**Technical specifications**
**Robot**
The telepresence robot used was RobAIR Social Touch. It was co-constructed with the fablab of the LIG (*Laboratoire d'Informatique de Grenoble*). It uses a ROS architecture, and the teleconferencing interface is based on WebRTC. Contrary to most of telepresence robots (Kristoffersson, Coradeschi & Loutfi, 2013), it does not look like a screen on wheels, but is closer to a slightly anthropomorphic robot carrying a tablet computer. On either side of the "head" of the robot, one omnidirectional microphone Behringer B5 is placed. Signals recorded by both microphones are digitalized by an audio interface UR22MKII (Steinberg), and sent to R's headphones in order to reproduce a pseudo binaural hearing. R's voice is emitted by a loudspeaker JBL GO+ which is placed under the tablet computer.

**Echo cancelation**
A local wireless router was used, the network latency was therefore negligible. However, visual and vocal signals cannot be transmitted instantaneously by WebRTC applications. As (Počta & Komperda, 2016), we obtained a mouth-to-ear delay of around 150 ms, which is pretty good according to telecommunication standards (ITU, 2003). This delay means, though, that there is an echo effect: the pilot R can hear her/his voice emitted by the loudspeakers of the robot and recorded by its microphones. This echo effect can greatly affect the quality of communication, so we used an algorithm of echo-cancelation. It consists simply in reducing the volume of R's headphones when s/he is talking, and waiting 150 ms before returning to the usual volume.

**Noise sources**
We used two loudspeakers JBL GO+ as sound sources. One was placed next to a coffee machine in the corridor adjacent to R's room. It diffused an amplified audio recording of the coffee machine startup (condition B). The other loudspeaker was placed in a room adjacent to L's room and diffused drill noises (condition D), which were pre-recorded with the microphones of the robot (condition C). Both sound sources were calibrated at 55 dB(A) with a sound level meter (Lutron SL-4001) placed at the location of the subjects. Moreover, the local space was particularly reverberant, with a reverberation time of about 0.8 s.

**Recordings and calibration**

The voices of both subjects were captured by two wireless headworn microphones Sennheiser HSP4 and digitalized with an audio interface UR22MKII (Steinberg). R's signal was sent to L through the loudspeakers of the robot. However, L's signal was only used for measurement purpose. What R heard of the local space was recorded by the microphones of the robot. In addition, the signal of the internal microphone of the computer, as well as the monitor signal of the stream heard by R were also recorded, in order to be able to track noise occurrences.

The headphones worn by R were an AKG K242. They were calibrated with an artificial ear (Brüel & Kjær), in order to make the sounds perceived by R as loud as those in L's room. The sound attenuation through these headphones is negligible. The loudspeaker of the robot was also calibrated with a sound level meter (Lutron SL-4001), to ensure that the loudness of R's voice was faithfully transmitted.

## Analysis of the results

14 groups of 2 subjects participated in this experiment. Most of them were native French speakers (25/28), two were fluent in French and one has a basic level, but sufficient to read the questions. Noise sequences were annotated and we extracted each keyword answered by subjects R (ex: "green"), and each pattern of questions read by subjects L (ex: "What is the color of…?"). We studied only the keywords / questions which were repeated at least 50 times among all the tests.

Recordings were filtered with a A-weighting digital filter (Zhivomirov, 2019) and voice intensity was computed by segments of 20 ms. Global results for maximum intensity can be seen on **Figure 2**. However, these results hide a great variability between subjects and between key-words / questions which could bias the analysis. Indeed, the number of each keyword in each condition varies from on experiment to another, because the answers of subjects R are not constrained. Therefore, in order to properly quantify the differences between each condition, we implemented a linear mixed effect model with *R*, following the tutorial of (Winter, 2013). This method allows to build a very simple model from the data, and at the same time, it provides measures of statistical significance. Written in *R* formula, this model is:

$$max\ intensity \sim noise\ condition\ +\ (1|keyword)\ +\ (1|subject)$$

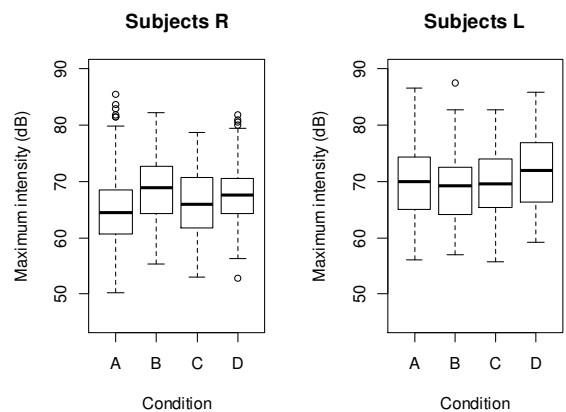

**Figure 2.** *Maximum intensity in each noise condition*

Summary of the models are shown in Table 1. They are coherent with the previous visualization, which means that we can base our observations on both figures. First, we note that the subjects R increased their maximum intensity in the 3 noisy conditions, while the subjects L spoke louder only in condition D, that is, when they could hear the noise. This shows that the Lombard Effect can be observed during telepresence robotics. However, it seems that for everyday noise with a moderate intensity level and when the subjects are focused on a question/answer task, this effect is very small, in the range of 1 to 3 dB. By way of comparison, (Winkworth Alison L. & Davis Pamela J., 1997) found an increase up to 10 dB SPL with a Cocktail party noise at 55 dB SPL for reading and monologue tasks.

|  | Subjects R | | | | Subjects L | | | |
| --- | --- | --- | --- | --- | --- | --- | --- | --- |
| **Condition** | **A** | **B** | **C** | **D** | **A** | **B** | **C** | **D** |
| **Maximum intensity (dB)** | 64.61 | + 2.98 | + 1.19 | + 2.38 | 69.19 | - 0.06 | - 0.30 | + 2.24 |
| **Standard error (dB)** | 1.11 | 0.36 | 0.37 | 0.36 | 1.86 | 0.21 | 0.21 | 0.20 |
| **Number of extracts** | 1263 | 160 | 146 | 150 | 1125 | 118 | 129 | 143 |
| **Statistical significance** | $\chi^2(1) = 98.47 \,;\, p < 2.2\, e^{-16}$ | | | | $\chi^2(1) = 132.59 \,;\, p < 2.2\, e^{-16}$ | | | |

**Table 1.** *Results of the linear mixed effect model (applied separately for R-data and L-data). Statistical significance was obtained by comparison with the null-model:* $max\ intensity \sim 1 + (1|keyword) + (1|subject)$

Another interesting result is that the increase of intensity for subjects R depends on the noise condition. It is higher in condition B vs C and D, namely when the noise was played in the same room as the pilot, and not in the headphones, which is the opposite of the results expected from (Garnier, Henrich & Dubois, 2010). However, the noises used in condition B were very different from the ones used in C/D. Indeed, they presented some intensity spikes up to 61 dB(A). Moreover, a posteriori recordings with the robot in R's room also showed that the intensity of the noises B was about 3 dB louder during stationary phase than the noises C/D. Besides, it is worth noticing that while the subjects R were talking, the volume of their headphones was reduced from 50% to 10%, in order to perform echo cancelation, which means that the intensity of noises C/D was greatly reduced at the time they spoke, which was not the case in condition B.

The most interesting result concerns the difference between condition C and D. The subjects R heard the same noises in both conditions, but their increase of intensity was greater in condition D, when the subjects L could also hear the noise and adapt to it. This may highlight an effect of entrainment: the subjects R increased their voice intensity not only because of the noise, but also because their interlocutors L were speaking louder. Such observations were also made by (Székely, Keane & Carson-Berndsen, 2015). However, this effect was not observed for the subjects L, who did not increase their voice intensity in condition B and C. This may be explained by the nature of their task (reading) or because they were not able to hear the noise which made their interlocutors speak louder.

## Discussion and conclusion

In order to study the Lombard effect in the context of telepresence robotics, we performed an experiment with pairs of subjects (R and L), focused on a question/answer task. Four noise conditions were tested:
A - without noise, B - only R hears the noise, C - only R hears the noise over headphones, and D - R and L hear the noise. The noise occurrences were very short and perceived as accidental by the participants. However, whenever they were able to hear the noise, they had a tendency to speak a bit louder. An entrainment effect was also noted for the subjects R, who spoke louder when L could also hear the noise. These increases of voice intensity being very subtle, they could easily be mistaken as expressive variations indicating socio-affects. Further work will involve comparing the pitch and durations of vocal productions in the four different noise conditions.

## Acknowledgements

We would like to thank Coriandre Villain (GIPSA-lab) for his technical assistance during the calibration. Thanks also to Emeline Le Goff and Zoé Giorgis (Univ. Grenoble Alpes) for their support during the development of the experimenter's interface and for annotating the data.